%% file: main.tex
\documentclass{sig-alternate-05-2015}

\usepackage{amsmath}
\usepackage{amssymb}
\usepackage{color}
\usepackage{subfigure}
\usepackage{url}
\usepackage{graphicx}
\usepackage{enumitem}
\usepackage[all]{nowidow}

\graphicspath{{graphics/}}

\usepackage{multirow}

\def\trace{\mathop{\sf tr}}

\def\0{\mathbf{0}}
\def\a{\mathbf{a}}
\def\b{\mathbf{b}}

\def\v{\mathbf{v}}
\def\x{\mathbf{x}}

\def\A{\mathbf{A}}
\def\B{\mathbf{B}}

\def\I{\mathbf{I}}

\def\L{\mathbf{L}}
\def\P{\mathbf{P}}

\def\S{\mathbf{S}}

\def\V{\mathbf{V}}
\def\W{\mathbf{W}}

\def\Y{\mathbf{Y}}
\def\Z{\mathbf{Z}}
\def\Lambda{\boldsymbol{\lambda}}

\def\Acal{\mathcal{A}}
\def\Ccal{\mathcal{C}}

\def\Pcal{\mathcal{P}}
\def\Ycal{\mathcal{Y}}
\def\Tcal{\mathcal{T}}

\definecolor{Red}{rgb}{1.0,0.0,0.0}

\let\oldbibliography\thebibliography
\renewcommand{\thebibliography}[1]{%
  \oldbibliography{#1}%
  \setlength{\itemsep}{-0.2pt}%
}

\title{Low-Rank Factorization of Determinantal Point Processes for
Recommendation}

\numberofauthors{3} 
\author{
\alignauthor
Mike Gartrell\\
       \affaddr{Microsoft}\\
       \email{mike.gartrell@acm.org}
\alignauthor
Ulrich Paquet\titlenote{Currently at Apple.} \\
       \affaddr{Microsoft} \\
       \email{ulripa@microsoft.com}
\alignauthor 
Noam Koenigstein\\
       \affaddr{Microsoft}\\
       \email{noamko@microsoft.com}
}

\makeatletter
\let\@copyrightspace\relax
\makeatother

\begin{document}

\maketitle

\input{abstract.tex} 

\input{intro.tex}

\input{model.tex}

\input{evaluation.tex}

\input{related-work.tex}

\input{conclusions.tex}

\input{acknowledgements.tex}

\bibliographystyle{abbrv}
\bibliography{paper}

\end{document}

%% file: abstract.tex
\begin{abstract}
Determinantal point processes (DPPs) have garnered attention as an elegant
probabilistic model of set diversity.  They are useful for a number of subset
selection tasks, including product recommendation.  DPPs are parametrized by a
positive semi-definite kernel matrix.  In this work we present a new method for
learning the DPP kernel from observed data using a low-rank factorization of
this kernel.  We show that this low-rank factorization enables a learning
algorithm that is nearly an order of magnitude faster than previous approaches,
while also providing for a method for computing product recommendation
predictions that is far faster (up to 20x faster or more for large item
catalogs) than previous techniques that involve a full-rank DPP kernel. 
Furthermore, we show that our method provides equivalent or sometimes better
predictive performance than prior full-rank DPP approaches, and better
performance than several other competing recommendation methods in many cases.
We conduct an extensive experimental evaluation using several real-world
datasets in the domain of product recommendation to demonstrate the utility of
our method, along with its limitations.
\end{abstract}

%% file: intro.tex

\section{Introduction}

Subset selection problems arise in a number of applications, including
recommendation~\cite{gillenwater2014EM}, document
summarization~\cite{kulesza2011learning, lin2012learning}, and Web
search~\cite{kulesza2011k}.
In these domains, we are concerned with selecting a good subset of high-quality
items that are distinct.  For example, a recommended subset of products
presented to a user should have high predicted ratings for that user while also
being diverse, so that we increase the chance of capturing the user's interest
with at least one of the recommended products.

Determinantal point processes (DPPs) offer an attractive model for such tasks,
since they jointly model \emph{set diversity} and \emph{quality or popularity},
while offering a compact parameterization and efficient algorithms for
performing inference. A distribution over sets that encourages diversity is of
particular interest when recommendations are complementary; for example, when a
shopping basket contains a laptop and a carrier bag, a complementary addition to
the basket would typically be a laptop cover, rather than another laptop.

DPPs can be parameterized by a $M \times M$ positive semi-definite $\L$ matrix,
where $M$ is the size of the item catalog.  There has been some work focused on
learning DPPs from observed data consisting of example
subsets~\cite{affandi2014learning, gillenwater2014EM, kulesza2011learning,
mariet15}, which is a challenging learning task that is conjectured to be
NP-hard~\cite{kulesza2012}. The most recent of this work has involved learning a
nonparametric full-rank $\L$ matrix~\cite{gillenwater2014EM, mariet15} that does
not constrain $\L$ to take a particular parametric form, which becomes
problematic with large item catalogs, as we will see in this paper.  In
contrast, we present a method for learning a low-rank factorization of $\L$,
which scales much better than full-rank approaches and in some cases provides
better predictive performance.  The scalability improvements allow us to
train our model on larger datasets that are infeasible with a full-rank
DPP, while also opening the door to computing online recommendations as
required for many real-world applications.

In addition to the applications mentioned above, DPPs have been used for a
variety of machine learning tasks~\cite{kang2013fast,
kulesza2010structured,
kulesza2012,
kwok2012priors,
snoek2013}.  We focus on the recommendation task of ``basket
completion'' in this work, where we compute predictions for the next item that
should be added to a shopping basket, given a set of items already present in
the basket.  This task is at the heart of online retail experiences, such as the
Microsoft Store.\footnote{www.microsoftstore.com}

Our work makes the following contributions:
\begin{enumerate}[itemsep=0pt]
  \item We present a low-rank DPP model, including algorithms for learning from
  observed data and computing predictions in the basket-completion scenario.
  \item We perform a detailed experimental evaluation of our model on several
  real-world datasets, and show that our approach scales substantially better
  than a full-rank DPP model, while providing equivalent or better predictive
  performance than the full-rank model.  We attribute our improvements in
  predictive performance to the novel use of regularization in our model.
  \item In addition to comparing our approach to a full-rank DPP, we also
  compare to several other models for basket completion and show significant
  improvements in predictive performance in many cases.
\end{enumerate}

Section~\ref{sec:model} gives a formal description of our model and describes
our algorithms for learning and prediction.  In Section~\ref{sec:evaluation} we
present a detailed evaluation of our model in terms of predictive performance,
and training and prediction run time.  We survey related work in
Section~\ref{sec:related-work}.

%% file: model.tex
\vspace{3.5em}
\section{Model}
\label{sec:model}
\subsection{Background}

A DPP is a distribution over configurations of points.\footnote{
DPPs originated in statistical mechanics~\cite{macchi1975}, where they were used
to model distributions of fermions.  Fermions are particles that obey the Pauli
exclusion principle, which indicates that no two fermions can occupy the same
quantum state.  As a result, systems of fermions exhibit a repulsion or
``anti-bunching'' effect, which is described by a DPP.   This repulsive behavior
is a key characteristic of DPPs, which makes them a capable model for diversity.
}
In this paper we deal only with discrete DPPs, which describe a distribution
over a discrete ground set of items $\Ycal = {1, 2, \ldots, M}$, which we also
call the item catalog.
A discrete DPP on $\Ycal$ is a probability measure $\Pcal$ on $2^{\Ycal}$ (the
power set or set of all subsets of $\Ycal$), such that for any $Y \subseteq
\Ycal$, the probability $\Pcal(Y)$ is specified by $\Pcal(Y) \propto
\det(\L_Y)$.  In the context of basket completion, $\Ycal$ is the item catalog
(inventory of items on sale), and $Y$ is the subset of items in a user's basket;
there are $2^{|\Ycal|}$ possible baskets.
The notation $\L_Y$ denotes the principal submatrix of the DPP kernel $\L$
indexed by the items in $Y$.  Intuitively, the diagonal entry $L_{ii}$ of the
kernel matrix $\L$ captures the importance or quality of item $i$, while the
off-diagonal entry $L_{ij} = L_{ji}$ measures the similarity between items $i$
and $j$.

The normalization constant for $\Pcal$ follows from the observation that
$\sum_{Y' \subseteq \Ycal} \det(\L_{Y'}) = \det(\L + \I)$.
The value $\det(\L_{Y})$ associates a ``volume'' to basket $Y$, and its
probability is normalized by the ``volumes'' of all possible baskets $Y'
\subseteq \Ycal$. Therefore, we have
\begin{align} \label{eq:dpp}
\Pcal(Y) = \frac{\det(\L_Y)}{\det(\L + \I)} \ .
\end{align}
We use a low-rank factorization of the $M \times M$ $\L$ matrix,
\begin{equation} \label{eq:Llowrank}
\L = \V \V^T \ ,
\end{equation}
for the $M \times K$ matrix $\V$, where $M$ is the number of items in the item
catalog and $K$ is the number of latent trait dimensions.  As we shall see in
this paper, this low-rank factorization of $\L$ leads to significant efficiency
improvements compared to a model that uses a full-rank $\L$ matrix when it comes
to model learning and computing predictions.  This also places an implicit
constraint on the space of subsets of $\Ycal$, since the model is restricted to
place zero probability mass on subsets with more than $K$ items
(all eigenvalues of $\L$ beyond $K$ are zero).
We see this from the observation that a sample from a DPP will not be larger
than the rank of $\L$~\cite{gillenwater2014approx}.

\subsection{Learning}

Our learning task is to fit a DPP kernel $\L$ based on a collection of $N$
observed subsets $\Acal = \{A_1, \ldots, A_n\}$ composed of items from the item
catalog $\Ycal$.  These observed subsets $\Acal$ constitute our training data,
and our task is to maximize the likelihood for data samples drawn from the same
distribution as $\Acal$.
The log-likelihood for seeing $\Acal$ is
\begin{align}
f(\V) & = \log \Pcal(\Acal | \V) = \sum_{n=1}^{N} \log \Pcal(\Acal_n | \V) \\
& = \sum_{n=1}^{N} \log \det(\L_{[n]}) - N \log \det(\L + \I)
\end{align}
where $[n]$ indexes the observations or objects in $\Acal$. We call the
log-likelihood function $f$, to avoid confusion with the matrix $\L$.
Recall from (\ref{eq:Llowrank}) that $\L = \V \V^T$.

The next two subsections describe how we perform optimization and
regularization for learning the DPP kernel.

\subsection{Optimization Algorithm}

We determine the $\V$ matrix by gradient ascent.
Therefore, we want to quickly
compute the derivative $\partial f / \partial \V$, which will be a $M \times K$
matrix. For $i \in 1, \ldots, M$ and $k \in 1, \ldots, K$, we need a matrix of
scalar derivatives, 
\[ \left\{ \frac{\partial f}{\partial \V} \right\}_{ik} =
\frac{\partial f}{\partial v_{ik}} \ .
\]
Taking the derivative of each term of the log-likelihood, we have 
\begin{align}
\frac{\partial f}{\partial v_{ik}}
& =
\sum_{n : i \in [n]} \frac{\partial}{\partial v_{ik}} \Big( \log \det(\L_{[n]})
\Big) - N \frac{\partial}{\partial v_{ik}} \Big( \log \det(\L + \I) \Big)
\nonumber
\\
& =
\sum_{n : i \in [n]} \trace \left( \L_{[n]}^{-1}  \frac{\partial \L_{[n]}}{\partial v_{ik}} \right) - N \trace \left( (\L + \I)^{-1}  \frac{\partial (\L + \I)}{\partial v_{ik}} \right) \ .
\end{align}
To compute the first term of the derivative, we see that 
\begin{align}
\trace \left( \L_{[n]}^{-1} \frac{\partial \L_{[n]}}{\partial v_{ik}} \right) =
\a_{i} \cdot \v_k + \sum_{j = 1}^{M} a_{ji} v_{jk} \ ,
\end{align}
where $\a_i$ denotes row $i$ of the matrix $\A = \L_{[n]}^{-1}$ and $\v_k$
denotes column $k$ of $\V_{[n]}$.
Note that $\L_{[n]} = \V_{[n]} \V_{[n]}^T$.
Computing $\A$ is a usually a relatively inexpensive operation, since the number
of items in each training instance $A_n$ is generally small for many
recommendation applications.

To compute the second term of the derivative, we see that  
\begin{align}
\trace \left( (\L + \I)^{-1}  \frac{\partial (\L + \I)}{\partial v_{ik}} \right)
= \b_{i} \cdot \v_k + \sum_{j = 1}^{M} b_{ji} v_{jk}
\end{align}
where $\b_i$ denotes row $i$ of the matrix $\B = \I_m - \V(\I_k + \V^T \V)^{-1}
\V^T$.  Computing $\B$ is a relatively inexpensive operation, since we are
inverting a $K \times K$ matrix with cost $O(K^3)$, and $K$ (the number of
latent trait dimensions) is usually set to a small value.

\subsubsection{Stochastic Gradient Ascent}

We implement stochastic gradient ascent with a form of momentum known as
Nesterov's Accelerated Gradient (NAG) \cite{Nesterov1983}:
\begin{align}
\W_{t + 1} &= \beta \W_t + (1 - \beta) * \epsilon \nabla f(\V_t + \beta \W_t) \\
\V_{t + 1} &= \V_t + \W_{t + 1}
\end{align}
where $\W$ accumulates the gradients, $\epsilon > 0$ is the learning rate, $\beta
\in [0, 1] $ is the momentum/NAG coefficient, and $\nabla f(\V + \beta \W_t)$ is
the gradient at $\V + \beta \W_t$.

We use the following schedule for annealing the learning rate:
\begin{align}
\epsilon_t &= \frac{\epsilon_0}{1 + t / T} 
\end{align}
where $\epsilon_0$ is the initial learning rate, $t$ is the iteration counter,
and $T$ is number of iterations for which $\epsilon$ should be kept nearly
constant.  This serves to keep $\epsilon$ nearly constant for the first $T$
training iterations, which allows the algorithm to find the general location of
the local maximum, and then anneals $\epsilon$ at a slow rate that is known from
theory to guarantee convergence to a local maximum~\cite{Robbins_Monro_1951}.
In practice, we set $T$ so that $\epsilon$ is held nearly fixed until the
iteration just before the test log-likelihood begins to decrease (which
indicates that we have likely ``jumped'' past the local maximum), and we find
that setting $\beta = 0.95$ and $\epsilon_0 = 1.0 \times 10 ^{-5}$ works well
for the datasets used in this paper.  Instead of computing the gradient using a
single training instance for each iteration, we compute the gradient using more
than one training instance, called a ``mini-batch''.  We find that a mini-batch
size of 1000 instances works well for the datasets we tested.

\subsection{Regularization}
\label{sec:regularization}

We add a quadratic regularization term to the log-like\-li\-hood, based on item
popularity, to discourage large parameter values and avoid overfitting.  Since not
all items in the item catalog are purchased with the same frequency, we encode
prior assumptions into the regularizer.  The motivation for using item
popularity in the regularizer is that the magnitude of the $K$-dimensional item
vector can be interpreted as the popularity of the item, as shown
in~\cite{gillenwater2014approx, kulesza2012}.
\begin{align}
f(\V) = \sum_{n=1}^{N} \log \det(\L_{[n]}) - N \log \det(\L + \I) -
\frac{\alpha}{2}\sum_{i=1}^{M} \lambda_{i} \|\v_{i}\|^2
\end{align}
where $\v_i$ is the row vector from $\V$ for item $i$, and $\lambda_i$ is an
element from a vector $\Lambda$ whose elements are inversely
proportional to item popularity,
\begin{align}
\Lambda &= \left( \frac{1}{C(1)}, \frac{1}{C(2)}, \ldots,
\frac{1}{C(i)} \right) \ ,
\end{align}
where $C(i)$ is the number of occurrences of item $i$ in the training data.

Taking the derivative of each term of the log-likelihood with this
regularization term, we now have
\begin{align}
\begin{split}
\frac{\partial f}{\partial v_{ik}}
& =
\sum_{n : i \in [n]} \frac{\partial}{\partial v_{ik}} \Big( \log \det(\L_{[n]})
\Big) - N \frac{\partial}{\partial v_{ik}} \Big( \log \det(\L + \I) \Big) \\
&\quad - \frac{\alpha}{2} \sum_{i=1}^{M} \lambda_{i} \frac{\partial}{\partial
v_{ik}} \Big( \|\v_{i}\|^2 \Big) \\
& = 
\sum_{n : i \in [n]} \trace \left( \L_{[n]}^{-1}  \frac{\partial
\L_{[n]}}{\partial v_{ik}} \right) - N \trace \left( (\L + \I)^{-1} 
\frac{\partial (\L + \I)}{\partial v_{ik}} \right) \\ 
&\quad - \alpha \lambda_{i}
v_{ik} \ .
\end{split}
\end{align}

\subsection{Predictions}

We seek to compute singleton next-item predictions, given a set of observed
items.  An example of this class of problem is ``basket completion'', where we
seek to compute predictions for the next item that should be added to shopping
basket, given a set of items already present in the basket. 

We use a $k$-DPP to compute next-item predictions.  A $k$-DPP is a distribution
over all subsets $Y \in \Ycal$ with cardinality $k$, where $\Ycal$ is the ground
set, or the set of all items in the item catalog.
Next item predictions are done via a conditional density.
We compute the probability of the observed basket $A$, consisting of $k$ items.
For each possible item to be recommended, given the basket,
the basket is enlarged with the new item to $k+1$ items.
For the new item, we determine the probability of the new 
set of $k+1$ items, given that $k$ items are already in the basket.
This machinery is also applicable when recommending a set $B$,
which may contain more than one added item, to the basket.

A $k$-DPP is obtained by
conditioning a standard DPP on the event that the set $Y$, a random set drawn
according to the DPP, has cardinality $k$.
Formally, for the $k$-DPP $\Pcal^k$ we have:
\begin{align}
\Pcal^k(Y) = \frac{\det(\L_Y)}{\sum_{|Y'| = k}{\det(\L_{Y'})}} 
\end{align}
where $|Y| = k$. Unlike (\ref{eq:dpp}), the normalizer sums \emph{only}
over sets that have cardinality $k$.

As shown in~\cite{kulesza2012}, we can condition a $k$-DPP on the event that all
of the elements in a set $A$ are observed.  We use $\L^A$ to denote the kernel matrix 
for this conditional $k$-DPP (the same notation is used for the conditional
kernel of the corresponding DPP, since the kernels are the same); we show in
Section~\ref{subsec:efficient-cond} how to efficiently compute this conditional
kernel.  For a set $B$ not intersecting with $A$, where $|A| + |B| = k$ we have:
\begin{align}
\Pcal^k(\Y = A \cup B | A \subseteq \Y)
& \propto
\Pcal^k_L(\Y = A \cup B) \\
& \propto
\Pcal(\Y = A \cup B) \\
& \propto
\det(\L^A_B) \\
& =
\frac{\det(\L^A_B)}{Z^A_{k - |A|}}
\label{eq:cond-k-DPP}
\end{align}
where here $B$ is a singleton set containing the possible next item for which we
would like to compute a predictive probability.  $\L^A_B$ denotes the principal
submatrix of $\L^A$ indexed by the items in $B$.

Ref.~\cite{kulesza2012} shows that the kernel matrix $\L^A$ for a conditional DPP is
\begin{equation}
\L^A = \left( \left[ (\L + \I_{\bar{A}})^{-1} \right]_{\bar{A}} \right)^{-1} -
\I
\label{eq:conditional-orig}
\end{equation}
where $\left[ (\L + \I_{\bar{A}})^{-1} \right]_{\bar{A}}$ is the restriction of
$(\L + \I_{\bar{A}})^{-1}$ to the rows and columns
indexed by elements in $\Ycal - A$, and $\I_{\bar{A}}$ is the matrix
with ones in the diagonal entries indexed by elements of $\Ycal - A$ and zeroes
everywhere else.

The normalization constant for Eq.~\ref{eq:cond-k-DPP} is
\begin{equation}
Z^A_{k - |A|} = \sum_{|Y'| = k - |A| \atop A \cap Y' = \emptyset}
\det(\L^A_{Y'}) \ ,
\end{equation}
where the sum runs over all sets $Y'$ of size $k - |A|$ that are disjoint from $A$. How can we compute it analytically?

We see from~\cite{kulesza2012} that 
\begin{equation}
Z_k = \sum_{|Y'| = k} \det(\L_{Y'}) = e_k(\lambda_1, \lambda_2, \ldots,
\lambda_M)
\end{equation}
where $\lambda_1, \allowbreak \lambda_2, \allowbreak \ldots, \allowbreak
\lambda_M$ are the eigenvalues of $\L$ and $e_k(\lambda_1, \allowbreak
\lambda_2, \allowbreak \ldots, \allowbreak \lambda_M)$ is the $k$th elementary
symmetric polynomial on $\lambda_1, \allowbreak \lambda_2, \allowbreak \ldots,
\allowbreak \lambda_M$.\footnote{Recall that when $\L = \V \V^T$ is defined in a low-rank form,
then all eigenvalues $\lambda_i = 0$ for $i > K$,
greatly simplifying the computation. When $\L$ is full rank, this is not the case.
Section \ref{sec:evaluation} compares 
the practical performance of a full-rank and low-rank $\L$.
}

Therefore, to compute the conditional probability for a single item $b$ in
singleton set $B$, given the appearance of items in a set $A$, we have
\begin{align}
\Pcal^k_L(\Y = A \cup B | A \subseteq \Y) 
& = 
\frac{\det(\L^A_B)}{Z^A_{k - |A|}} \\
& =
\frac{L^A_{bb}}{Z^A_1} \\
&=
\frac{L^A_{bb}}{e_1(\lambda^A_1, \lambda^A_2, \ldots, \lambda^A_N)}
\end{align}  
where $\lambda^A_1, \allowbreak \lambda^A_2, \allowbreak \ldots, \allowbreak
\lambda^A_N$ are the eigenvalues of $\L^A$ and $e_1(\lambda^A_1, \allowbreak
\lambda^A_2, \allowbreak \ldots, \allowbreak \lambda^A_N)$ is the first
elementary symmetric polynomial on these eigenvalues.

\subsubsection{Efficient DPP Conditioning}
\label{subsec:efficient-cond}

The conditional probability used for prediction (and hence set recommendation or basket completion)
uses $\L^A$ in Eq.~\ref{eq:conditional-orig},
which requires
two inversions of large matrices.
These are expensive operations, particularly
for a large item catalog (large $M$). In this section we describe a
way to efficiently condition the DPP $\L$ kernel that is enabled by our low-rank
factorization of $\L$. 

Ref.~\cite{gillenwater2014approx} shows that for a DPP with kernel $\L$, the
conditional kernel $\L^A$ with minors satisfying
\begin{equation}
\Pcal(\Y = Y \cup A | A \subseteq \Y) = \frac{\det(\L^A_Y)}{\det(\L^A + \I)}
\label{eq:conditional}
\end{equation}
on $Y \subseteq \Ycal \setminus A$, can be computed from $\L$ by the rank-$|A|$
update
\begin{equation}
\L^A = \L_{\bar{A}} - \L_{\bar{A},A} \L_{A}^{-1} \L_{A,\bar{A}}
\label{eq:conditional-rank-update}
\end{equation}
where $\L_{\bar{A},A}$ consists of the $|\bar{A}|$ rows and $A$ columns of $\L$.
Substituting $\V$ into Eq.~\ref{eq:conditional-rank-update} gives
\begin{align}
\L^A 
& =
\V_{\bar{A}} \Z^A \V_{\bar{A}}^T \label{eq:LZ}
\end{align}
where 
\begin{equation}
\Z^A = \I - \V_A^T (\V_A \V_A^T)^{-1} \V_A
\label{eq:z-matrix} \ .
\end{equation}
$\Z^A$ is a projection matrix, and is thus idempotent: $\Z^A = (\Z^A)^2$.
Since $\Z^A$ is also symmetric, we have $\Z^A = (\Z^A)^T$,
and substituting $\Z^A = \Z^A (\Z^A)^T$ into (\ref{eq:LZ}) yields
\begin{align}
\L^A 
& =
\V_{\bar{A}} \Z^A (\Z^A)^T \V_{\bar{A}}^T \\
& =
\V^A (\V^A)^T \label{eq:conditional-low-rank}
\end{align}
where
\begin{equation}
\V^A = \V_{\bar{A}} \Z^A \ .
\end{equation}
Conditioning the DPP using Eq.~\ref{eq:conditional-low-rank} requires
computing the inverse of a $|A| \times |A|$ matrix, as shown in
Eq.~\ref{eq:z-matrix}, which is $O(|A|^3)$.
This is much less expensive than the matrix inversions in
Eq.~\ref{eq:conditional-orig} when $|A| \ll M$, which we expect for most
recommendation applications.  For example, in online shopping applications, the
size of a shopping basket ($|A|$) is generally far smaller than the size of the
item catalog ($M$).

%% file: evaluation.tex

\section{Evaluation}
\label{sec:evaluation}
In this section we compare the low-rank DPP model with a full-rank DPP that uses
a fixed-point optimization algorithm called Picard iteration~\cite{mariet15} for
learning. We wish to showcase the advantage of low-rank DPPs in practical
scenarios such as basket completion. First, we compare test log-likelihood of
low-rank and full-rank DPPs and show that the low-rank model's ability to
generalize is comparable to that of the full-rank version. We also compare the
training times and prediction times of both algorithms and show a clear
advantage for the low-rank model presented in this paper.  Our implementations
of the low-rank and full-rank DPP models are written in Julia, and we perform
all experiments on a Windows 10 system with 32 GB of RAM and an Intel Core
i7-4770 CPU @ 3.4 GHz.

Comparing test log-likelihood values and training time is consistent with
previous studies~\cite{gillenwater2014EM,mariet15}. Log-likelihood values
however are not always correlated with other evaluation metrics. In the
recommender systems community it is usually more accepted to use other metrics
such as precision@$k$ and mean percentile rank (MPR). In this paper we also
compare DPPs (low-rank and full-rank) to other competing methods using
these more ``traditional'' evaluation metrics.

Our experiments are based on several datasets:
\begin{enumerate}[leftmargin=*]
	\item \textbf{Amazon Baby Registries -} 
This public dataset consists of 111,006 registries of baby products
from 15 different categories (such as ``feeding'', ``diapers'', ``toys'', etc.),
where the item catalog and registries for each category are disjoint.
The public dataset was obtained by collecting baby registries from
\url{amazon.com} and was used by previous DPP
studies~\cite{gillenwater2014EM,mariet15}.
In particular,~\cite{gillenwater2014EM} provides an in-depth description of this
dataset.  To maintain consistency with prior work, we used a random split of
70\% of the data for training and 30\% for testing.  We use $K = 30$ trait
dimensions for the low-rank DPP models trained on this data.  While the Baby
Registries dataset is relatively large, previous studies analyzed each of its
categories separately. We maintain this approach for the sake of consistency
with prior work.
	
We also construct a dataset composed of the concatenation of the three most
popular categories: apparel, diaper, and feeding.  This three-category dataset
allows us to simulate data that could be observed for department stores that
offer a wide range of items in different product categories.  Its construction
is deliberate, and concatenates three disjoint subgraphs of basket-item purchase
patterns. This dataset serves to highlight differences between DPPs and models
based on matrix factorization (MF), as there are no items or baskets shared
between the three subgraphs.
Collaborative filtering-based MF models -- which model each basket and item with
a latent vector -- will perform poorly for this dataset, as the latent vectors
of baskets and items in one subgraph could be arbitrarily rotated, without
affecting the likelihood or predictive error in any of the other subgraphs.  MF
models are invariant to global rotations of the embedded vectors. However, for
the concatenated dataset, these models are also invariant to arbitrary rotations
of vectors in each disjoint subgraph for the concatenated data set, as there are
no shared observations between the three categories. A global ranking based on
inner products could then be arbitrarily affected by the basket and item
embeddings arising from each subgraph.

The low-rank approximation presented in this paper facilitates scaling-up DPPs
to much larger datasets. Therefore, we conducted experiments on two
additional real-world datasets, as we explain next.

	\item \textbf{MS Store -}			
This is a proprietary dataset composed of shopping baskets purchased in
Microsoft's Web-based store~\url{microsoftstore.com}. It consists of 243,147
purchased baskets composed of 2097 different hardware and software items.  We
use a random split of 80\% of the data for training and 20\% for testing. 
For the low-rank DPP model trained on this data, we use $K = 15$ trait
dimensions.

	\item \textbf{Belgian Retail Supermarket -}
This is a public dataset~\cite{brijs99, brijs2003} composed of
shopping baskets purchased over three non-consecutive time periods from a
Belgian reatil supermarket store.  There are 88,163 purchased baskets, composed
of 16,470 unique items.  We use a random split of 80\% of the data for training
and 20\% for testing.  We use $K = 76$ trait dimensions for the low-rank DPP
model trained on this data.
\end{enumerate}

Since we are interested in the basket completion task, which requires baskets
containing at least two items, we remove all baskets containing only one item
from each dataset before splitting the data into training and test sets.

We determine convergence during training of both the low-rank and full-rank DPP
models using
\begin{align}
\frac{ |f(\V_{t + 1}) - f(\V_t))| }{ | f(\V_t)| } \leq \delta \nonumber
\end{align}
which measures the relative change in training log-likelihoods from one
iteration to the next.  We set $\delta = 1.0 \times 10^{-5}$.  

\subsection{Full Rank vs.~Low Rank}
\input{tab_log-likelihood}
We begin with comparing test log-likelihood values of the low-rank DPP model
presented in this paper with the full-rank DPP trained using Picard iteration.
Table~\ref{tab:loglikelihoods} depicts the average test log-likelihoods values
of both models across the different categories of the Baby Registries dataset as
well as the MS Store dataset. In the Baby Registry dataset the full-rank model
seems to perform better in 9 categories compared with 6 categories for the
low-rank model, and for the MS Store dataset the full-rank model performed
better. The differences in the log-likelihood values are small, and as we show
in Section~\ref{sec:more_metrics} these differences do not necessarily translate
into better results for other evaluation metrics.

\subsubsection{Training Time}
\begin{figure}[tb]
	\centering
	\subfigure[Training time, in seconds]{
		\includegraphics[width=0.48\textwidth]{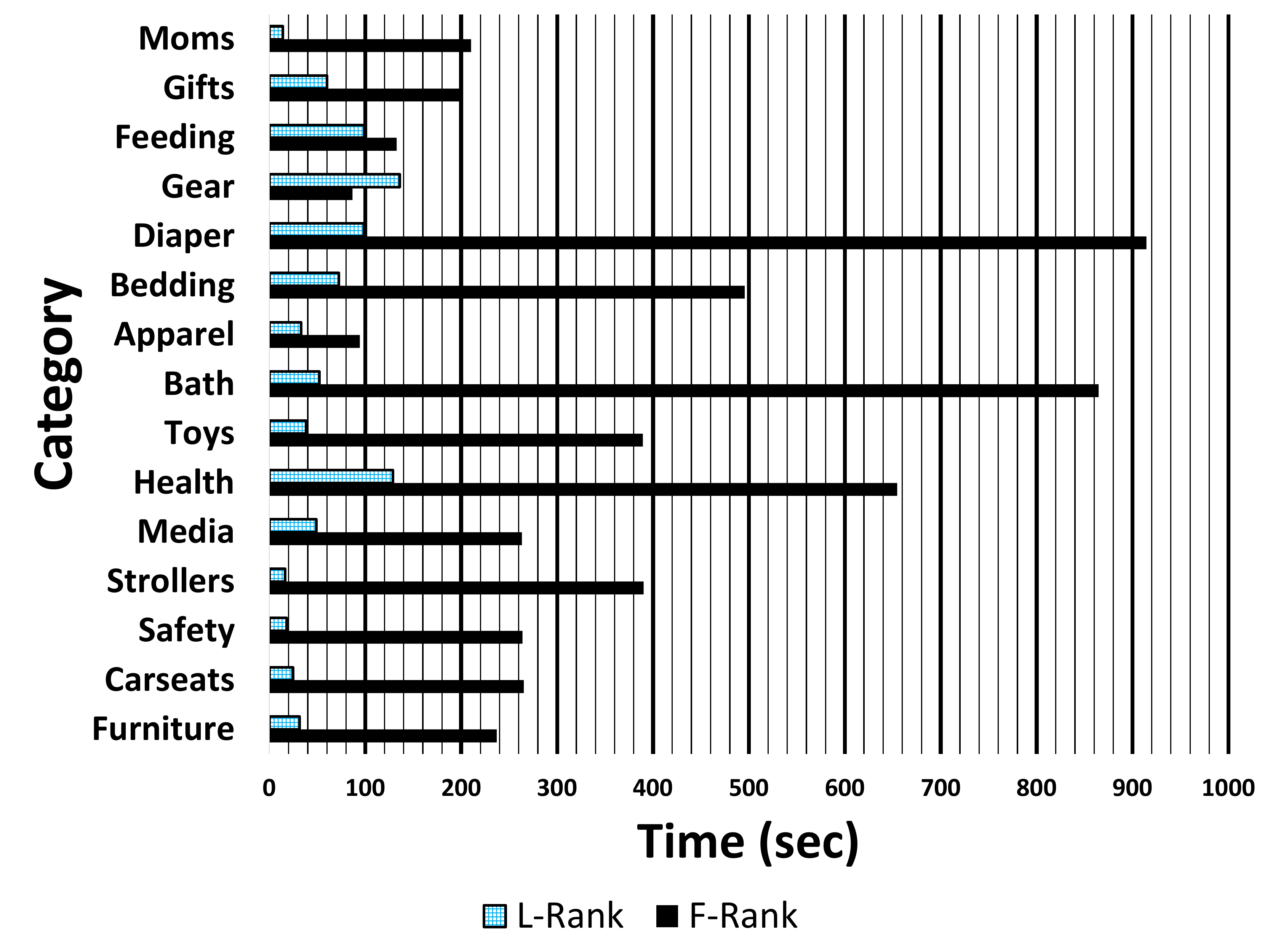}
		\label{fig:training_time}            
	}
	\subfigure[Average Prediction time per basket, in milliseconds]{          
		\includegraphics[width=0.48\textwidth]{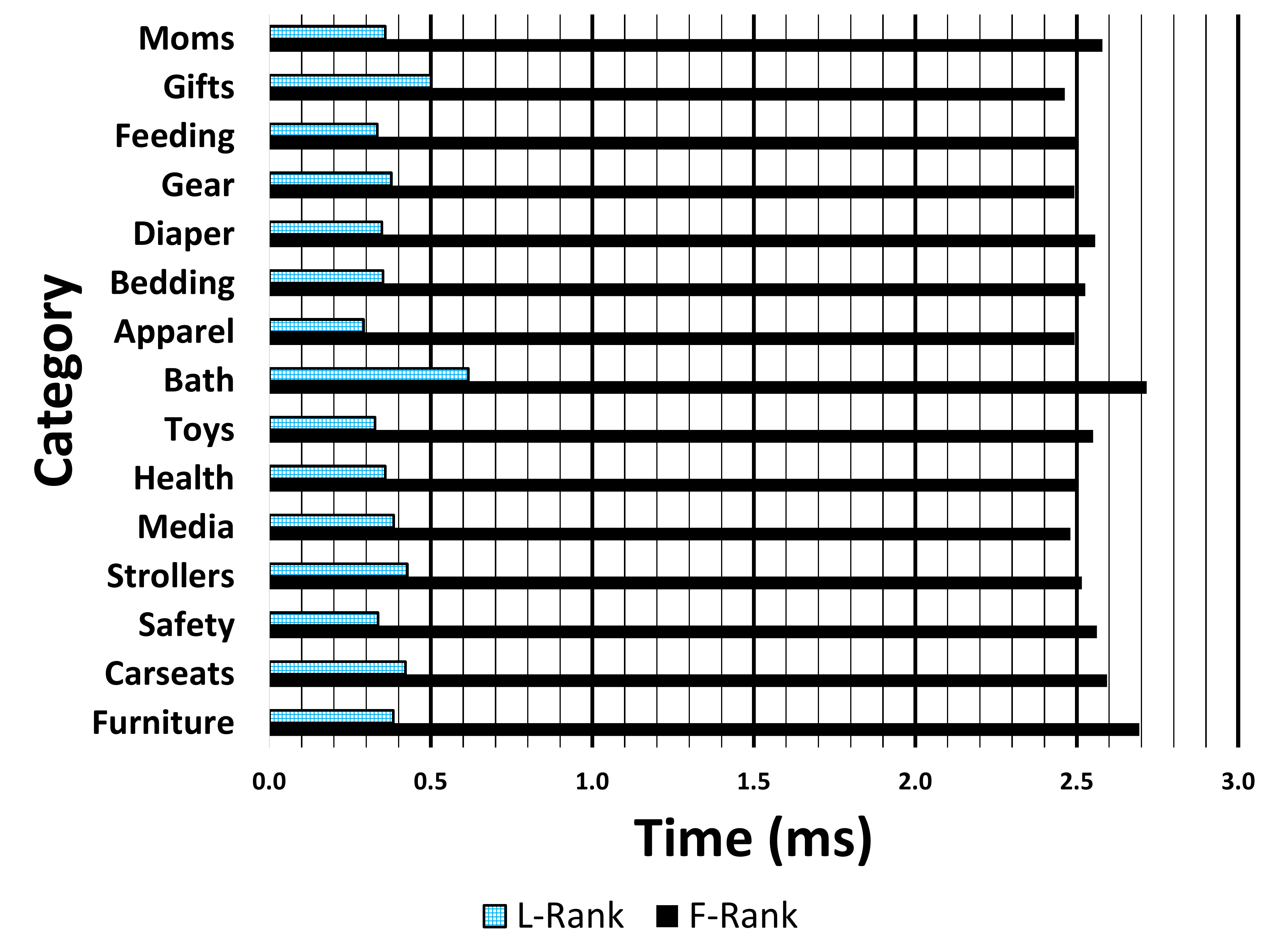}
		\label{fig:prediction_time}
	}
	\vspace{-1.5em}
	\caption{Training and prediction time of low rank DPP (L-Rank) vs.~full rank DPP (F-Rank).}
	\label{fig:time}
	\vspace{-1.0em}
\end{figure}
A key contribution of the Picard iteration method was the improvement of
training time (convergence time) by up to an order of magnitude~\cite{mariet15}
compared to previous methods. However, the Picard iteration method requires
inverting an $M \times M$ full-rank $(\L +\I)$ matrix, where $M$ is the number
of items in the catalog. This matrix inversion operation has a $O(M^3)$ time
complexity. In the low-rank model, this operation is replaced by an inversion of
a $K \times K$ matrix where $K \ll M$, and training is performed by stochastic
gradient ascent. This translates into considerably faster training times,
particularity in cases where the item catalog is large.

Figure~\ref{fig:training_time} depicts the training time in seconds of the
full-rank (F-Rank) model vs. the low-rank (L-Rank) DPP model described in this
paper.  Table~\ref{tab:num-iterations} shows the number of iterations required
for each model to reach convergence. Training times are shown for each of the 15
categories in the Baby Registry dataset. In all but one category, the training
time of the low-rank model was considerably faster.  On average, the low-rank
model is 8.9 times faster to train than the full-rank model.

\begin{table}
  \centering
  \begin{tabular}{|c|c|c|}
  \hline
  \textbf{Category} & \textbf{L-Rank} & \textbf{F-Rank} \\ \hline
  Mom & 67 & 1294 \\ \hline
  Gifts & 126 & 1388 \\ \hline
  Feeding & 68 & 123 \\ \hline
  Gear & 82 & 136 \\ \hline
  Diaper & 83 & 1065 \\ \hline
  Bedding & 88 & 772 \\ \hline
  Apparel & 48 & 129 \\ \hline
  Bath & 64 & 1664 \\ \hline
  Toys & 66 & 970 \\ \hline
  Health & 68 & 1337 \\ \hline
  Media & 126 & 958 \\ \hline
  Strollers & 53 & 1637 \\ \hline
  Safety & 59 & 1306 \\ \hline
  Carseats & 54 & 1218 \\ \hline
  Furniture & 54 & 1277 \\ \hline   	
  \end{tabular}
  \vspace{-0.5em}
  \caption{Number of training iterations to reach convergence, for low-rank DPP
  (L-Rank) and full-rank DPP (F-Rank) models}
  \label{tab:num-iterations}
  \vspace{-1.0em}
\end{table}

\subsubsection{Prediction Time}
In production settings, training is usually performed in advance (offline),
while predictions are computed per request (online).
A typical real-world recommender system models at least thousands of items (and
often much more). The ``relationships'' between items changes slowly with time
and it is reasonable to train a model once a day or even once a week.
The number of possible baskets, however, is vast and depends on the number of
items in the catalog. Therefore, it is wasteful and sometimes impossible to
pre-compute all possible basket recommendations in advance.
The preferred choice in most cases would be to compute predictions online, in
real time.

High prediction times may overload online servers, leading to high response
times and even failure to provide recommendations.  The ability to compute
recommendations efficiently is key to any real-world recommender system. Hence,
in real-world scenarios prediction times are usually much more important than
training times.

Previous DPP studies~\cite{gillenwater2014EM, mariet15} focused on training
times and did not offer any improvement in prediction times. In fact, as we show
next, the average prediction time spikes for the full-rank DPP when the size of
the item catalog reaches several thousand, and quickly becomes impractical in
real-world settings where the inventory of items is large and fast online
predictions are required. Our low-rank model facilitates far faster prediction
times and scales well for large item catalogs, which is key to any practical use
of DPPs. We believe this contribution opens the door to large-scale use
of DPP models in commercial settings.

In Figure~\ref{fig:prediction_time} we compare the average prediction time for a
test-set basket for each of the 15 categories in the Baby Registry dataset.
This figure shows the average time to compute predictive probabilities for all
possible items that could added to the basket for a given test basket instance,
where the set of possible items are those items found in the item catalog but
not in the test basket.  Since the catalog is composed of a maximum of only 100
items for each Baby Registry category, due to way that the dataset was
constructed, we see that these prediction times are quite small. Again we notice
a clear advantage for the low-rank model across all categories:
the average prediction time for the full-rank model is 2.55 ms per basket,
compared with 0.39 ms for the low-rank model (6.8 times faster). Since number of
items in the catalog for each baby registry category is small (100 items), we
also measured the prediction time for the MS Store dataset, which contains 2,097
items. Due to the much larger item catalog, the average time per a single basket
prediction increases significantly to 1.66 seconds, which is probably too slow
for many real-world recommender systems. On the other hand, the average
prediction time of the low-rank model depends mostly on the number of trait
dimensions in the model and takes only 83.6 ms per basket on average. These
numbers indicate a speedup factor of 19.9.

Our low-rank DPP model also provides substantial savings in memory consumption
as compared to the full-rank DPP.  For example, the MS Store dataset, composed
of a catalog of 2097 items, would require $2097 \times 2097 \times 8
 \text{ bytes} = 35.18 \text{ MB}$ to store the full-rank DPP kernel matrix
(assuming 64-bit floating point numbers), while only $2097 \times 15 \times 8
\text{ bytes} = 251.6 \text{ KB}$ would be required to store the low-rank $\V$
matrix with $K = 15$ trait dimensions. Therefore, the low-rank model requires
approximately 140 times less memory to store the model parameters in this
example, and this savings increases with larger item catalogs.

\begin{figure*}[t!] 
	\centering
	\includegraphics[width=0.90\textwidth]{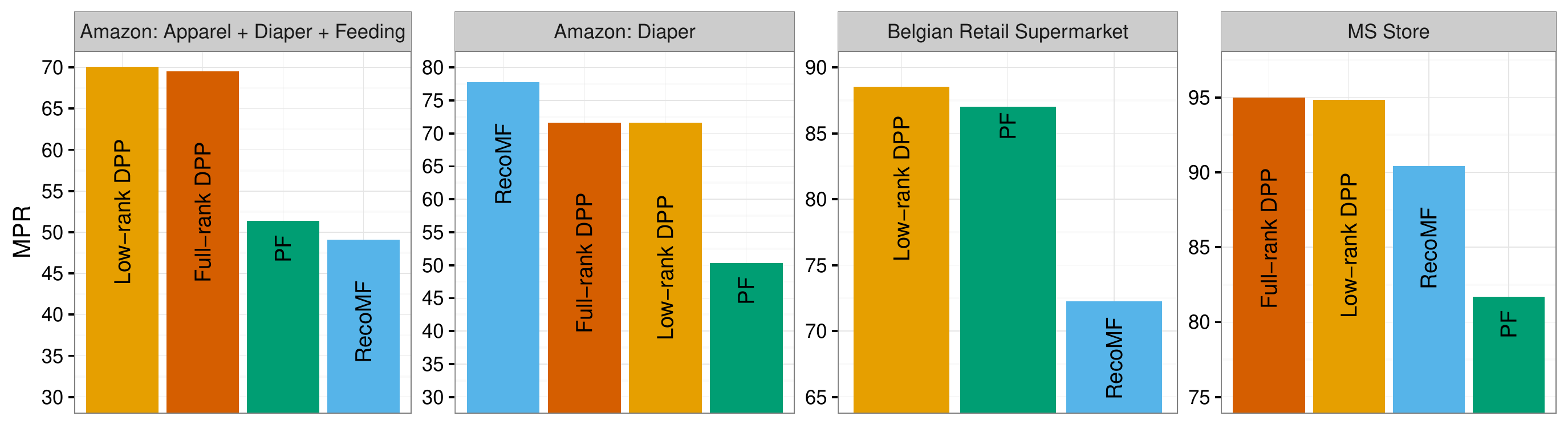}
	\vspace{-1.0em}
	\caption{Mean Percentile Rank (MPR)}
	\label{fig:MPR}
\end{figure*}

\begin{figure*}[t!] 
	\centering
	\includegraphics[width=0.90\textwidth]{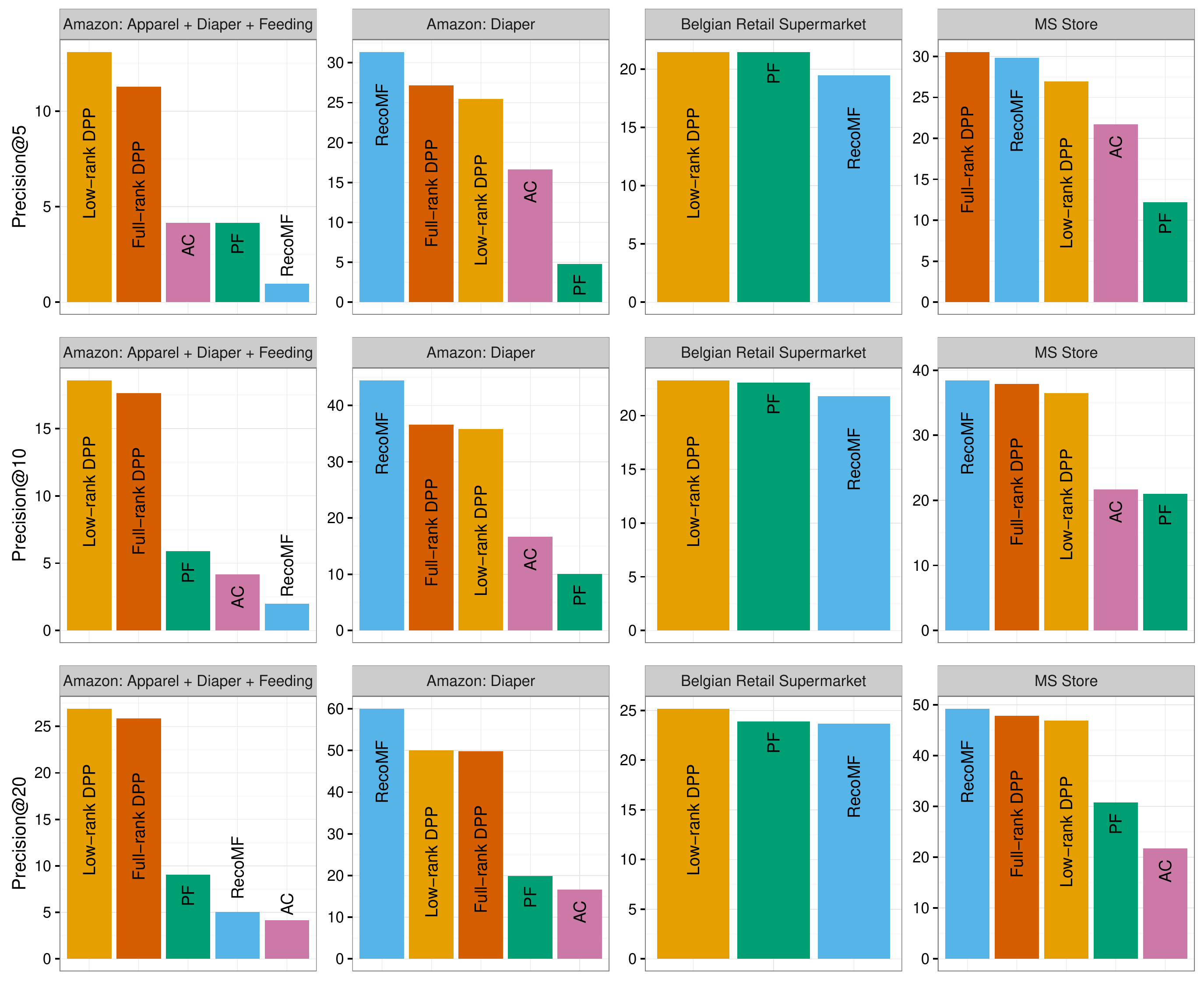}
	\vspace{-1.0em}
	\caption{Precision@$k$}
	\label{fig:precisionAtK}
	\vspace{-0.5em}
\end{figure*}

\begin{figure*}[t!] 
	\centering
	\includegraphics[width=0.90\textwidth]{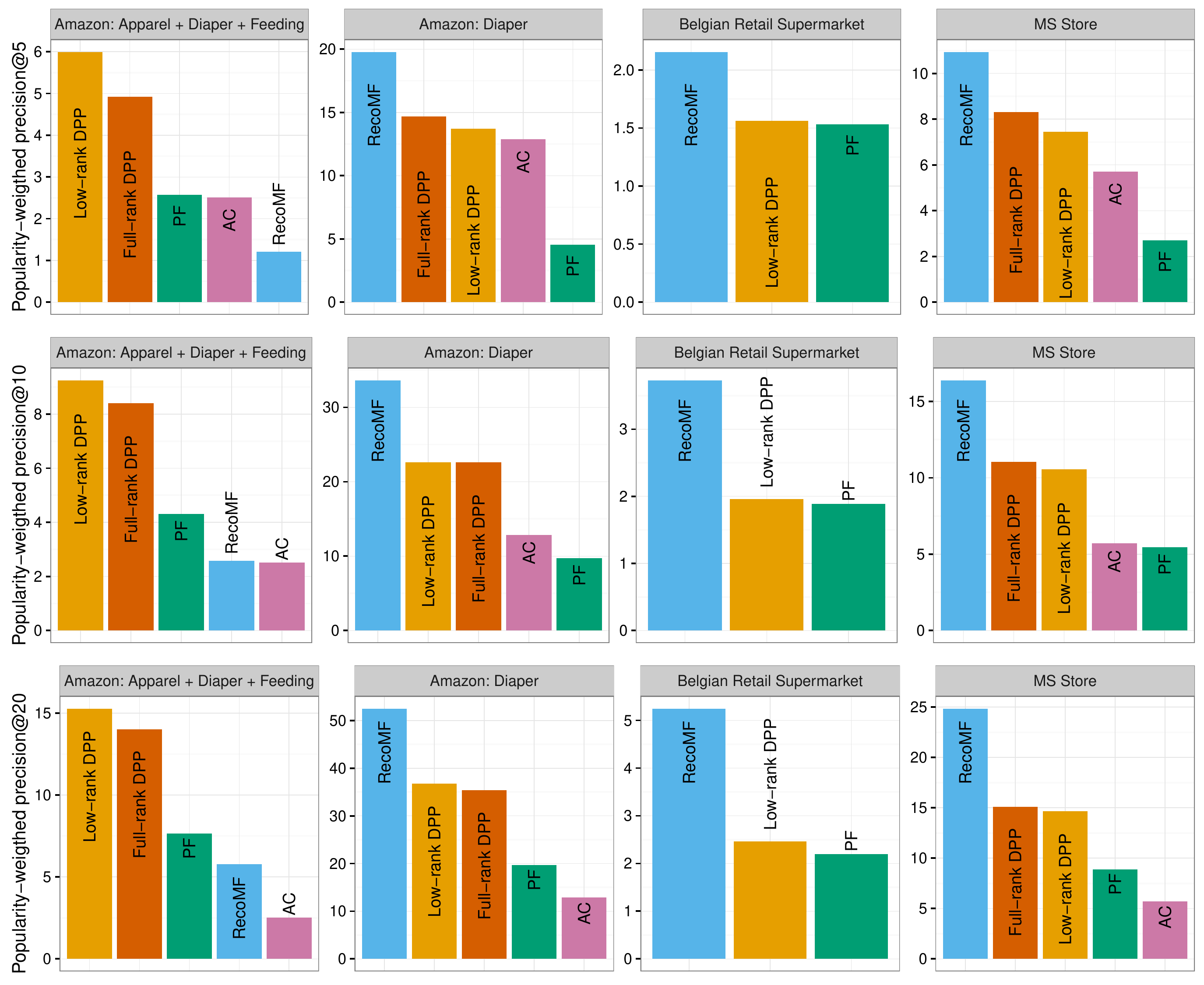}
	\vspace{-1.0em}
	\caption{Popularity-weighted precision@$k$.  These results show a limitation
	of the DPP models.  Since this metric biases precision@$k$ towards less popular
	items, we see that the RecoMF model often provides better predictive
	performance for less popular items.}
	\label{fig:popPrecisionAtK}
	\vspace{-0.5em}
\end{figure*}

\subsection{Basket Completion and Recommendations}
\label{sec:more_metrics}
Previous papers have evaluated DPP recommendations by comparing test
log-likelihood values. In this section we also consider more
``traditional'' evaluation metrics commonly used in the recommender systems
community.

We formulate the basket-completion task as follows.  Let $Y_n$ be a subset of
$n\geq2$ co-purchased items (i.e, a basket) from the test-set.  In order to
evaluate the basket completion task, we pick an item $i \in Y_n$ at random and
remove it from $Y_n$. We denote the remaining set as $Y_{n-1}$. Formally,
$Y_{n-1} = Y_n \diagdown \, \{i\}$. Given a ground set of possible items
$\Ycal=1,2,...,M$, we define the candidates set $\Ccal$ as the set of all items
except those already in $Y_{n-1}$; i.e., $\Ccal = \Ycal \diagdown \,Y_{n-1} $.
Our goal is to identify the missing item $i$ from all other items in $\Ccal$.

We compare the low-rank DPP model with the full-rank DPP model. We also consider
several other competing models for the basket completion task:
\begin{enumerate}[leftmargin=*]
	\item \textbf{Poisson Factorization (PF) - } 
Poisson factorization (PF)~\cite{gopalan2015} is a recent variant of
probabilistic matrix factorization that has been shown to work well with
implicit recommendation data, such as clicks or purchases.  PF models user-item
interactions, such as clicks or purchases, with factorized Poisson
distributions, and learns sparse, non-negative trait vectors for latent user
preferences and item attributes in a low-dimensional space.  Gamma priors are
placed on the trait vectors; we set the gamma shape and rate hyperparameters to
0.3, following~\cite{chaney2015, gopalan2015}.  The PF model is not sensitive to
these settings, as indicated in~\cite{chaney2015, gopalan2015}.  We use a
publicly available implementation of PF~\cite{PFimpl}.
 (Note that~\cite{PFimpl} is actually an implementation of PF with a social
component; we disable the social component for our tests, resulting in a model
equivalent to PF, since our data does not involve a social graph).
	\item \textbf{Reco Matrix Factorization (RecoMF) - }
RecoMF is a matrix factorization model~\cite{paquet2013one} that is used as the
recommendation system for Xbox Live.  Sigmoid functions are used to
model the odds of a user liking or disliking an item, and RecoMF learns latent
trait vectors for users and items, along with user and item biases.  Unlike PF,
RecoMF requires the generation of synthetic negative training instances, and
uses a scheme for sampling negatives based on popularity.  RecoMF places
Gaussian priors on the trait vectors, and gamma hyperpriors on each.  We use the
hyperparameter settings described in~\cite{paquet2013one}, which have been found
to provide good performance for implicit recommendation data.
	\item \textbf{Associative Classifier (AC) - } 
We use an associative classifier as a competing method, since association rules
are often used for market basket
analysis~\cite{agrawal1993mining,kotsiantis2006}.  Our associative classifier is
the publicly available implementation~\cite{CBAimpl} of the Classification Based
on Associations (CBA) algorithm~\cite{ma1998integrating}.  We use minimum
support and minimum confidence thresholds of 1.0\% and 20.0\%, respectively.
Since associative classifiers don't provide probability estimates for all
possible sets, the model therefore cannot compute rankings for all of the
candidate items in $\Ccal$, and we therefore cannot reasonably compute MPR.
\end{enumerate}

The matrix-factorization models are parameterized in terms of users and items. 
Since we have no explicit users in our data, we construct ``virtual'' users from
the contents of each basket for the purposes of our evaluation, where a new user
$u_m$ is constructed for each basket $b_m$.  Therefore, the set of items that
$u_m$ has purchased is simply the contents of $b_m$.  Additionally, we use $K =
40$ trait dimensions for the matrix-factorization models.

In the following evaluation we consider three measures:
\begin{enumerate}[leftmargin=*]
	\item \textbf{Mean Percentile Rank (MPR) - }
Computing the Percentile Rank of an item requires the ability to rank the item
$j$ against all other items in $\Ccal$. Therefore, the MPR evaluation results
don't include the AC model, which ranks only those items for which an
association rule was found.  For DPPs and other competing methods we ranked the
items according to their probabilities to complete the missing set $Y_{n-1}$.
Namely, given an item $i$ from the candidates set $\Ccal$, we denote by $p_i$
the probability $P\Large(Y_n \cup \{i\} |Y_{n-1}\Large)$.
The Percentile Rank (PR) of the missing item $j$ is defined by
\begin{equation}
\text{PR}_j = \frac{\sum_{j'\in \Ccal}\mathbb{I}\Large( p_j \ge
p_{j'}\Large)}{|\Ccal|} \times 100\% \nonumber
\end{equation}
where $\mathbb{I}\Large( \cdot \Large)$ is an indicator function and $|\Ccal|$
is the number of items in the candidates set.    The Mean Percentile Rank (MPR) is
the average PR of all the instances in the test-set:
\begin{equation}
\text{MPR} = \frac{\sum_{t \in \Tcal} \text{PR}_t}{|\Tcal|} \nonumber
\end{equation}  
where $\Tcal$ is the set of test instances.  MPR is a recall-oriented metric
commonly used in studies that involve implicit recommendation data~\cite{hu2008,
li2010}.  $\text{MPR} = 100$ always places the held-out item for the test
instance at the head of the ranked list of predictions, while $\text{MPR} = 50$
is equivalent to random selection.
	\item \textbf{Precision@$k$ - }
We define precision@$k$ as 
\begin{equation}
\text{precision@}k = \frac{\sum_{t \in \Tcal} \mathbb{I}[\text{rank}_t \leq
k]}{|\Tcal|}
\nonumber
\end{equation}
where $\text{rank}_t$ is the predicted rank of the held-out item for test
instance $t$.  In other words, precision@$k$ is the fraction of instances in the
test set for which the predicted rank of the held-out item falls within the top
$k$ predictions.
	\item \textbf{Popularity-weighted precision@$k$ - }
Datasets used to evaluate recommendation systems typically
contain a popularity bias~\cite{steck2011}, where users are more likely to
provide feedback on popular items.  Due to this popularity bias, conventional
metrics such as MPR and precision@$k$ are typically biased toward popular items.
Using ideas from~\cite{steck2011}, we propose popularity-weighted precision@$k$:
\begin{align}
\text{popularity-weighted precision@}k = \nonumber \\ 
\frac{\sum_{t \in \Tcal} w_t \mathbb{I}[\text{rank}_t \leq k]}{\sum_{t \in
\Tcal} w_t} \nonumber
\end{align}
where $w_t$ is the weight assigned to the held-out item for test instance $t$,
defined as
\begin{equation}
w_i \propto \frac{1}{C(t)^\beta} \nonumber
\end{equation}
where $C(t)$ is the number of occurrences of the held-out item for test instance
$t$ in the training data, and $\beta \in [0, 1]$.  The weights are normalized,
so that $\sum_{j \in \Ycal} w_j = 1$.  This popularity-weighted precision@$k$
measure assumes that item popularity follows a power-law.  By assigning more
weight to less popular items, for $\beta > 0$, this measure serves to bias
precision@$k$ towards less popular items.  For $\beta = 0$, we obtain the
conventional precision@$k$ measure.  We set $\beta = 0.5$ in our evaluation.
\end{enumerate}

Figures~\ref{fig:MPR},~\ref{fig:precisionAtK}, and~\ref{fig:popPrecisionAtK}
show the performance of each method and dataset for our evaluation measures.
Note that we could not feasibly train the full-rank DPP or AC models on the
Belgian dataset, since these models do not scale to datasets with large item
catalogs.  The performance of low-rank and full-rank DPP models are generally
comparable on all models and metrics, with the low-rank DPP providing better
performance in some cases.  We attribute this advantage to the use of
regularization (an informative prior, from a Bayesian perspective) in our
low-rank model.  We see that the RecoMF model outperforms all other models on
all metrics for the Amazon Diaper dataset.  For all other datasets, the low-rank
DPP model outperforms on MPR by a sizeable margin, and is the only model to
consistently provide high MPR across all datasets.  For the precision@$k$
metrics, the low-rank DPP often leads, or provides good performance that is
close to the leader.  

We see interesting results for the Amazon apparel + diaper + feeding dataset. 
Surprisingly, the PF and RecoMF models provide a MPR of approximately 50\%, which
is equivalent to basket completion by random selection.  Recall that each category
in the Amazon baby registry dataset is disjoint.  Due to the formulation of the
likelihood function for models based on matrix factorization, these models learn
an embedding of item trait vectors that is mixed together across each disjoint
category.  This behavior results in the model mixing the predictions across each
category, e.g. recommending an item from category $A$ for a basket in category
$B$, which is never observed in the data, thus leading to degenerate results. 
We empirically observe that the DPP models do not have this issue, and are able to
effectively learn an embedding of items in this scenario: notice that the DPP
models provide an MPR of approximately 70\% for both the Amazon three-category
and single-category (diaper) datasets. 

\subsubsection{Limitations}
We include the popularity-weighted precision@$k$ results in
Figure~\ref{fig:popPrecisionAtK} to highlight a limitation of the DPP models.
For this metric RecoMF generally provides the best performance, with the DPP
models in second place.  As discussed in~\cite{paquet2013one}, this behavior may
result from the scheme for sampling negatives by popularity in RecoMF, which
tends to improve recommendations for less popular items.  We conjecture that a
different regularization scheme for our low-rank DPP model, or a Bayesian
version of this model that provides more robust regularization, may improve our
performance on this metric.  It is also important to note the limitations of
this metric, including the assumption that item popularity follows a power-law,
and the power-law exponent $\beta$ setting of 0.5 used when computing the metric
for each dataset.  Due to these limitations, the popularity-weighted
precision@$k$ results we present here may not fully reflect the empirical
popularity bias actually present in the data.

%% file: tab_log-likelihood.tex

\begin{table}
\begin{center}
\begin{small}
  \begin{tabular}{ | l | c | c | }    	
	
	\multicolumn{3}{c}{\textbf{Baby Registry}} \\
	\hline
	\textbf{Category} & \textbf{F-Rank} & \textbf{L-Rank} \\ \hline
      Furniture & -7.07391 & \textbf{-7.00022} \\ \hline
      Carseats & \textbf{-7.20197} & -7.27515\\ \hline
      Safety & -7.08845 & \textbf{-7.01632} \\ \hline
      Strollers & \textbf{-7.83098} & -7.83201\\ \hline
      Media & \textbf{-12.29392} & -12.39054\\ \hline
      Health & \textbf{-10.09915} & -10.36373 \\ \hline
	  Toys & \textbf{-11.06298} & -11.07322 \\ \hline
	  Bath & -11.89129 & \textbf{-11.88259} \\ \hline
	  Apparel & \textbf{-13.84652}  & -13.85295 \\ \hline
	  Bedding & \textbf{-11.53302} & -11.58239\\ \hline
	  Diaper &  \textbf{-13.13087} & -13.16574 \\ \hline
	  Gear & -12.30525 & \textbf{-12.17447} \\ \hline
	  Feeding & -14.91297 & \textbf{-14.87305}\\ \hline
	  Gifts & \textbf{-4.94114} & -4.96162 \\ \hline
	  Moms & -5.39681 & \textbf{-5.34985} \\ \hline
	  
	  	\multicolumn{3}{c}{}\\
	\multicolumn{3}{c}{\textbf{MS Store}} \\	\hline
	\textbf{} & \textbf{F-Rank} & \textbf{L-Rank} \\ 
	\hline
	All Products & \textbf{-15.10} & -15.23 \\ \hline
	  
   \end{tabular}
  \end{small}
  \end{center}
  \vspace{-1.5em}
  \caption{Average test log-likelihoods values of low-rank (L-Rank) and full-rank (F-Rank) DPPs.} 
  \vspace{-0.5em}
  \label{tab:loglikelihoods}
\end{table}

%% file: related-work.tex

\section{Related Work}
\label{sec:related-work}

Several learning algorithms for estimating the full-rank DPP kernel matrix from
observed data have been proposed.  Ref.~\cite{gillenwater2014EM}
presented one of the first methods for learning a non-parametric form of the DPP
kernel matrix, which involves an expectation-maximization (EM) algorithm.  This
work also considers using projected gradient ascent on the DPP log-likelihood
function, but finds that this is not a viable approach since it usually results
in degenerate estimates due to the projection step. 

In~\cite{mariet15}, a fixed-point optimization algorithm for DPP learning is
described, called Picard iteration. Picard iteration has the advantage of being
simple to implement and performing much faster than EM during training.  We show
in this paper that our low-rank learning approach is far faster than Picard
iteration and therefore EM during training, and that our low-rank representation
of the DPP kernel allows us to compute predictions much faster than any method
that uses the full-rank kernel.

Ref.~\cite{affandi2014learning} presented Bayesian methods for
learning a DPP kernel, with particular parametric forms for the similarity and
quality components of the kernel.  Markov chain Monte Carlo (MCMC) methods are
used for sampling from the posterior distribution over kernel parameters.  In
contrast to this work, and similar to~\cite{gillenwater2014EM, mariet15}, our
approach uses a non-parametric form of the kernel and therefore does not assume
any particular parametrization.

A method for partially learning the DPP kernel is studied
in~\cite{kulesza2011learning}.  The similarity component of the DPP kernel is
fixed, and a parametric from of the function for the quality component of the
kernel is learned.  This is a convex optimization problem, unlike the task of
learning the full kernel, which is a more challenging non-convex optimization
problem.

We focus on the prediction task of ``basket completion'' in this work, as it is
at the heart of the online retail experience.
For the purposes of evaluating our model, we compute predictions for the next
item that should be added to a shopping basket, given a set of items already
present in the basket. A number of approaches to this problem have been
proposed. Ref.~\cite{mild2003} describes a user-neighborhood-based collaborative
filtering method, which uses rating data in the form of binary purchases to
compute the similarity between users, and then generates a purchase prediction
for a user and item by computing a weighted average of the binary ratings for
that item.  A technique that uses logistic regression to predict if a user will
purchase an item based on binary purchase scores obtained from market basket
data is described in~\cite{lee2005}.  Additionally, other collaborative
filtering approaches could be applied to the basket completion problem, such
as~\cite{paquet2013one}, which is a one-class matrix factorization model.

%% file: conclusions.tex
\section{Conclusions}
\label{sec:conclusions}

In this paper we have presented a new method for learning the DPP kernel from
observed data, which exploits the unique properties of a low-rank factorization
of this kernel.  Previous approaches have focused on learning a full-rank
kernel, which does not scale for large item catalogs due to high memory
consumption and expensive operations required during training and when computing
predictions.  We have shown that our low-rank DPP model is substantially faster
and more memory efficient than previous approaches for both training and
prediction.  Furthermore, through an experimental evaluation using several
real-world datasets in the domain of recommendations for shopping baskets, we
have shown that our method provides equivalent or sometimes better predictive
performance than prior full-rank DPP approaches, while in many cases also
providing better predictive performance than competing methods.

%% file: acknowledgements.tex
\section{Acknowledgements}
\label{sec:acknowledgements}

We thank Gal Lavee and Shay Ben Elazar for many helpful discussions.  We thank
Nir Nice for supporting this work.

%% file: main.bbl
\begin{thebibliography}{10}

\bibitem{affandi2014learning}
R.~H. Affandi, E.~Fox, R.~Adams, and B.~Taskar.
\newblock Learning the parameters of determinantal point process kernels.
\newblock In {\em ICML}, pages 1224--1232, 2014.

\bibitem{agrawal1993mining}
R.~Agrawal, T.~Imieli{\'n}ski, and A.~Swami.
\newblock Mining association rules between sets of items in large databases.
\newblock In {\em Proc. of SIGMOD 1993}, pages 207--216, 1993.

\bibitem{brijs2003}
T.~Brijs.
\newblock Retail market basket data set.
\newblock In {\em Workshop on Frequent Itemset Mining Implementations
  (FIMI'03)}, 2003.

\bibitem{brijs99}
T.~Brijs, G.~Swinnen, K.~Vanhoof, and G.~Wets.
\newblock Using association rules for product assortment decisions: A case
  study.
\newblock In {\em KDD}, pages 254--260, 1999.

\bibitem{PFimpl}
A.~J. Chaney.
\newblock Social {P}oisson factorization ({SPF}).
\newblock \url{https://github.com/ajbc/spf}, 2105.

\bibitem{chaney2015}
A.~J. Chaney, D.~M. Blei, and T.~Eliassi-Rad.
\newblock A probabilistic model for using social networks in personalized item
  recommendation.
\newblock In {\em RecSys}, pages 43--50, 2015.

\bibitem{CBAimpl}
F.~Coenen.
\newblock {TLUCS} {KDD} implementation of {CBA} (classification based on
  associations).
\newblock \url{http://www.csc.liv.ac.uk/~frans/KDD/Software/CMAR/cba.html},
  2004.
\newblock Department of Computer Science, The University of Liverpool, UK.

\bibitem{gillenwater2014approx}
J.~Gillenwater.
\newblock {\em Approximate inference for determinantal point processes}.
\newblock PhD thesis, University of Pennsylvania, 2014.

\bibitem{gillenwater2014EM}
J.~A. Gillenwater, A.~Kulesza, E.~Fox, and B.~Taskar.
\newblock Expectation-maximization for learning determinantal point processes.
\newblock In {\em NIPS}, pages 3149--3157, 2014.

\bibitem{gopalan2015}
P.~Gopalan, J.~M. Hofman, and D.~M. Blei.
\newblock Scalable recommendation with hierarchical {P}oisson factorization.
\newblock In {\em UAI}, 2015.

\bibitem{hu2008}
Y.~Hu, Y.~Koren, and C.~Volinsky.
\newblock Collaborative filtering for implicit feedback datasets.
\newblock In {\em ICDM}, pages 263--272, 2008.

\bibitem{kang2013fast}
B.~Kang.
\newblock Fast determinantal point process sampling with application to
  clustering.
\newblock In {\em NIPS}, pages 2319--2327, 2013.

\bibitem{kotsiantis2006}
S.~Kotsiantis and D.~Kanellopoulos.
\newblock Association rules mining: A recent overview.
\newblock {\em GESTS International Transactions on Computer Science and
  Engineering}, 32(1):71--82, 2006.

\bibitem{kulesza2010structured}
A.~Kulesza and B.~Taskar.
\newblock Structured determinantal point processes.
\newblock In {\em NIPS}, pages 1171--1179, 2010.

\bibitem{kulesza2011k}
A.~Kulesza and B.~Taskar.
\newblock k-{DPP}s: Fixed-size determinantal point processes.
\newblock In {\em ICML}, pages 1193--1200, 2011.

\bibitem{kulesza2011learning}
A.~Kulesza and B.~Taskar.
\newblock Learning determinantal point processes.
\newblock In {\em UAI}, 2011.

\bibitem{kulesza2012}
A.~Kulesza and B.~Taskar.
\newblock Determinantal point processes for machine learning.
\newblock {\em Foundations and Trends in Machine Learning}, 5(2-3):123--286,
  2012.

\bibitem{kwok2012priors}
J.~T. Kwok and R.~P. Adams.
\newblock Priors for diversity in generative latent variable models.
\newblock In {\em NIPS}, pages 2996--3004, 2012.

\bibitem{lee2005}
J.-S. Lee, C.-H. Jun, J.~Lee, and S.~Kim.
\newblock Classification-based collaborative filtering using market basket
  data.
\newblock {\em Expert Systems with Applications}, 29(3):700--704, 2005.

\bibitem{li2010}
Y.~Li, J.~Hu, C.~Zhai, and Y.~Chen.
\newblock Improving one-class collaborative filtering by incorporating rich
  user information.
\newblock In {\em CIKM}, pages 959--968, 2010.

\bibitem{lin2012learning}
H.~Lin and J.~Bilmes.
\newblock Learning mixtures of submodular shells with application to document
  summarization.
\newblock In {\em UAI}, 2012.

\bibitem{ma1998integrating}
B.~Liu, W.~Hsu, and Y.~Ma.
\newblock Integrating classification and association rule mining.
\newblock In {\em KDD}, 1998.

\bibitem{macchi1975}
O.~Macchi.
\newblock The coincidence approach to stochastic point processes.
\newblock {\em Advances in Applied Probability}, pages 83--122, 1975.

\bibitem{mariet15}
Z.~Mariet and S.~Sra.
\newblock Fixed-point algorithms for learning determinantal point processes.
\newblock In {\em ICML}, pages 2389--2397, 2015.

\bibitem{mild2003}
A.~Mild and T.~Reutterer.
\newblock An improved collaborative filtering approach for predicting
  cross-category purchases based on binary market basket data.
\newblock {\em Journal of Retailing and Consumer Services}, 10(3):123--133,
  2003.

\bibitem{Nesterov1983}
Y.~Nesterov.
\newblock A method of solving a convex program- ming problem with convergence
  rate {O}(1/sqr(k)).
\newblock {\em Soviet Mathematics Doklady}, 27:327--376, 1983.

\bibitem{paquet2013one}
U.~Paquet and N.~Koenigstein.
\newblock One-class collaborative filtering with random graphs.
\newblock In {\em WWW}, pages 999--1008, 2013.

\bibitem{Robbins_Monro_1951}
H.~Robbins and S.~Monro.
\newblock A stochastic approximation method.
\newblock {\em The Annals of Mathematical Statistics}, 22(3):400--407, 1951.

\bibitem{snoek2013}
J.~Snoek, R.~Zemel, and R.~P. Adams.
\newblock A determinantal point process latent variable model for inhibition in
  neural spiking data.
\newblock In {\em NIPS}, pages 1932--1940, 2013.

\bibitem{steck2011}
H.~Steck.
\newblock Item popularity and recommendation accuracy.
\newblock In {\em RecSys}, pages 125--132, 2011.

\end{thebibliography}
